%% file: main.tex
\documentclass[11pt, a4paper, copyright, nonumbering]{skywork}
\usepackage[authoryear, sort&compress, round]{natbib}
\usepackage{dblfloatfix}
\usepackage{ulem}
\usepackage{caption}
\usepackage{dramatist}
\usepackage{xspace}
\usepackage{pifont}
\usepackage{multirow}
\usepackage{adjustbox}
\usepackage{tcolorbox}
\usepackage{xltabular}
\usepackage{longtable}
\usepackage{hyperref}
\usepackage{makecell}
\usepackage{algorithm}
\usepackage[noend]{algpseudocode}
\usepackage{amsfonts}
\usepackage{amsmath}
\usepackage{amssymb}
\usepackage{lineno}
\usepackage[bottom]{footmisc}
\usepackage{CJKutf8}
\usepackage{subfigure}
\usepackage{setspace}
\usepackage{titlesec}
\usepackage{enumitem}
\usepackage[framemethod=TikZ]{mdframed}
\usepackage{lipsum}
\usepackage{amsmath}
\usepackage{colortbl} 
\usepackage{booktabs} 
\usepackage[english]{babel}
\usepackage{caption}
\usepackage{subcaption}
\usepackage{epigraph}
\usepackage{wrapfig}
\usepackage{tablefootnote}
\usepackage{float}
\usepackage{dashrule}
\usepackage{arydshln}
\usepackage{footnote}
\usepackage{cleveref}
\usepackage{makecell}

\definecolor{greyrow}{rgb}{0.9, 0.9, 0.9} 
\definecolor{orangerow}{rgb}{1, 0.95, 0.9}
\definecolor{greenrow}{rgb}{0.9, 1, 0.9}
\definecolor{bluerow}{rgb}{0.9, 0.9, 1}
\definecolor{redrow}{rgb}{1, 0.9, 0.9}

\makeatletter
\def\@BTrule[#1]{%
  \ifx\longtable\undefined
    \let\@BTswitch\@BTnormal
  \else\ifx\hline\LT@hline
    \nobreak
    \let\@BTswitch\@BLTrule
  \else
     \let\@BTswitch\@BTnormal
  \fi\fi
  \global\@thisrulewidth=#1\relax
  \ifnum\@thisruleclass=\tw@\vskip\@aboverulesep\else
  \ifnum\@lastruleclass=\z@\vskip\@aboverulesep\else
  \ifnum\@lastruleclass=\@ne\vskip\doublerulesep\fi\fi\fi
  \@BTswitch}
\makeatother

\addto\extrasenglish{
}

 {\begin{list}{}%
         {\setlength{\leftmargin}{#1}}%
         \item[]%
 }
 {\end{list}}
 
\reportnumber{001} 

\renewcommand{\today}{}

\vspace{-5mm}
\title{\centering Skywork-Reward: Bag of Tricks for Reward Modeling in LLMs}
\author[*]{
\hspace{3em}
Chris Yuhao Liu$^*$, Liang Zeng$^*$, Jiacai Liu, Rui Yan, Jujie He, Chaojie Wang, \newline Shuicheng Yan, Yang Liu, Yahui Zhou 

\hspace{1em}
\small$^*$Equal contribution, \{yuhao.liuu, liang.zeng\}@kunlun-inc.com

Skywork AI, Kunlun Inc.
}
\correspondingauthor{}

\input{commands.tex}

\input{sections/abstract}

\begin{document}
\begin{CJK*}{UTF8}{gbsn}

\maketitle

\input{sections/introduction}
\input{sections/related}
\input{sections/method}
\input{sections/experiment}
\input{sections/conclusion}
\newpage
\bibliography{main}

\end{CJK*}
\end{document}

%% file: commands.tex
\renewcommand{\phi}{\varphi}

\renewcommand{\epsilon}{\varepsilon}
\renewcommand{\imath}{\mathrm{i}}

\newlength{\restsubwidth}
\newlength{\restsubheight}
\newlength{\restsubmoreheight}
\newcommand{\rest}[2]{%
        \settowidth{\restsubwidth}{\ensuremath{#2}}
        \settoheight{\restsubheight}{\ensuremath{{}_{#2}}}
        \ensuremath{{#1\hskip 0.5pt}_{\vrule\kern2pt\parbox[b][%
        4pt][b]{\the\restsubwidth}{%
                        \ensuremath{{}_{#2}}}}}
        }

%% file: sections/abstract.tex
\begin{abstract}
In this report,  we introduce a collection of methods to enhance reward modeling for LLMs, focusing specifically on data-centric techniques. We propose effective data selection and filtering strategies for curating high-quality open-source preference datasets, culminating in the Skywork-Reward data collection, which contains only 80K preference pairs---significantly smaller than existing datasets. Using this curated dataset, we developed the Skywork-Reward model series—Skywork-Reward-Gemma-27B and Skywork-Reward-Llama-3.1-8B—with the former currently holding the top position on the RewardBench leaderboard. Notably, our techniques and datasets have directly enhanced the performance of many top-ranked models on RewardBench, highlighting the practical impact of our contributions in real-world preference learning applications \footnote{The models and datasets are publicly available at \url{https://huggingface.co/collections/Skywork/skywork-reward-model-66d7fbdebae0e60d00a6b60d} and \url{https://huggingface.co/collections/Skywork/skywork-reward-data-collection-66d7fda6a5098dc77035336d}}. 
\end{abstract}

%% file: sections/introduction.tex
\section{Introduction}
Large language models~(LLMs) have achieved unprecedented success, demonstrating capabilities that were previously unattainable in both scope and performance \citep{gemma_2024, dubey2024llama, achiam2023gpt, team2023gemini, team2024gemini, team2024gemma}. This rapid advancement has fueled extensive research into aligning LLM outputs with user preferences \citep{bai2022training}. Among the various alignment strategies, reward modeling has emerged as a prominent and scalable approach for capturing these preferences \citep{lambert2024rewardbench, wang2024helpsteer2}. Reward models are explicitly trained to evaluate how well the LLM outputs align with the intended responses desired by users, effectively acting as evaluators during both fine-tuning and deployment \citep{dong2024rlhf, wang2023helpsteer, cai2024internlm2, wang2024arithmetic, wang2024direct}. 

Despite its potential, training reward models poses several significant challenges \citep{lambert2024rewardbench}, primarily due to the inherent complexity and variability of human preferences, which are difficult to represent exhaustively \citep{sanderson2010user}. 
Prior research has sought to address these challenges by improving model architectures \citep{ArmoRM, wang2024arithmetic} and developing customized loss functions \citep{cai2024internlm2, winata2024metametrics, lou2024uncertainty}, enabling reward models to better differentiate between nuanced preference pairs. These methods enhance the models' capacity to prioritize preferred responses while minimizing rejected ones, thereby improving alignment with user preferences. 
In addition to these efforts, the availability and quality of preference data play a pivotal role in the success of reward modeling. Unfortunately, open-source preference datasets are often noisy, with differences between preferred and rejected responses either overly subtle or inconsistently labeled \citep{wang2024helpsteer2, xu2024magpie, park2024offsetbias}. Such inconsistencies can significantly degrade the performance of reward models, underscoring the importance of meticulous data selection and filtering to ensure robust and reliable modeling.

In this paper, we propose a comprehensive suite of techniques to enhance reward modeling in LLMs, with a particular focus on the curation of high-quality preference data. Specifically, we introduce lightweight yet effective preference data collections, relying solely on publicly available sources to ensure transparency and reproducibility. Our data selection and filtering strategies are designed to prioritize preference pairs that contribute most effectively to improving model performance. Additionally, we conduct extensive ablation studies on various loss functions, focusing on optimizing the margin between preferred and rejected responses. Our experimental results demonstrate that the vanilla Bradley-Terry loss \citep{bradley1952rank,ouyang2022training} consistently outperforms alternative approaches, underscoring its robustness in reward modeling tasks.

We collectively employ these advanced training techniques to develop the Skywork-Reward model series and rigorously validate their effectiveness on the RewardBench benchmark \citep{lambert2024rewardbench}, demonstrating significant performance improvements with our proposed training techniques.
As of October 2024, the Skywork-Reward model series holds the first and seventh positions on the RewardBench leaderboard \citep{lambert2024rewardbench}. Furthermore, our curated Skywork-Reward preference data collection has been widely adopted in subsequent research efforts \citep{winata2024metametrics, lou2024uncertainty, yang2024regularizing,zhang2024general}, highlighting its value and applicability. To promote further research and innovation in reward modeling for LLMs, we publicly release both the Skywork-Reward model series and the corresponding preference data collection. 
We hope that these contributions will inspire the future development of more aligned and human-centered LLMs.

%% file: sections/related.tex
\section{Related Work}
Recent advancements in applying reinforcement learning techniques \citep{schulman2017proximal}, particularly Reinforcement Learning from Human Feedback (RLHF) \citep{bai2022training, casper2023open}, have shown substantial potential for enhancing LLMs. A key component of RLHF is the development of reward models \citep{dubey2024llama, gemma_2024, gao2023scaling}, which learn a reward function based on human preferences or task-specific objectives to guide LLMs toward desired behaviors. As discussed by \citet{lambert2024rewardbench}, reward modeling techniques can be broadly categorized into three categories based on the underlying model types: discriminative models, generative models, and implicit reward models through Direct Preference Optimization~(DPO). We briefly describe each of them as follows.

\paragraph{Discriminative Models} 
Discriminative reward models are commonly trained using the Bradley-Terry~(BT) \citep{bradley1952rank} loss, which aims to maximize the reward difference between pairwise comparisons—specifically, between chosen responses and rejected responses. These models estimate the probability that a given response is preferred over an alternative, making them well-suited for binary ranking tasks. While the core BT loss remains a standard component, considerable research has focused on enhancing data quality and refining the modeling framework. For example, the InternLM2-Reward models \citep{cai2024internlm2}, trained on 2.4 million human-annotated and AI-generated preference samples, are optimized to classify pairwise comparisons, ensuring a careful balance between helpfulness and harmlessness. \citet{yang2024regularizing} improve the generalization ability of reward models by introducing regularization in the hidden states, mitigating the risk of over-optimization on specific reward functions. In a complementary effort, \citet{park2024offsetbias} address inherent biases in reward models—such as the tendency to favor longer responses—by proposing de-biasing strategies in dataset construction. To capture more nuanced and complex preferences, models like Nemotron-Reward \citep{wang2024helpsteer2} leverage multi-dimensional reward signals, allowing for a more granular understanding of user preferences. Other methods introduce architectural modifications to boost performance. For instance, other than multi-dimensional rewards, ArmoRM \citep{ArmoRM, wang2024arithmetic} also utilizes a gating network that adaptively selects the most relevant reward dimension based on contextual information. Similarly, \citet{zhang2024general} explore the use of latent spaces within LLMs to model preferences, relying on similarity scores between responses to inform preference-based decisions. These advancements collaboratively push the boundaries of discriminative reward modeling, improving the ability of LLMs to align with diverse and subtle human preferences.

\paragraph{Generative Models} 
While discriminative models are widely adopted, generative models offer an alternative approach by directly using LLM-generated outputs to evaluate preference data \citep{zheng2023judging}. Generative models excel in providing nuanced, interpretable assessments, capturing subtle differences in language use, and offering deeper insights into the decision-making process. However, their performance in reward modeling tasks often lags behind discriminative models \citep{lambert2024rewardbench}, as they are not specifically optimized to rank or select between pairwise comparisons. To bridge this gap, \citet{wang2024direct} introduces an auxiliary task—response deduction—to enhance generative models' ability to judge pairwise comparisons effectively based on textual outputs. Similarly, Self-Taught \citep{wang2024self} improves generative models through contrastive learning \citep{khosla2020supervised}, enabling them to generate preference judgments without relying on human annotations.  Additionally, state-of-the-art chat-based LLMs like Gemini \citep{team2023gemini} and GPT-4o \citep{achiam2023gpt} demonstrate the potential of generative models by directly producing textual rewards. These advanced models leverage their powerful generative abilities to showcase the versatility of generative reward modeling in complex scenarios.

\paragraph{Implicit Rewards via DPO Models} 
A third category, Direct Preference Optimization~(DPO) \cite{rafailov2024direct}, enables RLHF without the requirement of an explicitly trained reward model. Instead, DPO derives a reward signal directly from the current policy and an initial supervised fine-tuned policy \citep{rafailov2024r}, effectively reparameterizing preference learning within the model itself. While DPO models are not able to assign reward signals like a discriminative model or a generative model in nature, implicit rewards can be computed when a corresponding supervised fine-tuned version of the model is available \citep{bellagente2024stable, ivison2023camels}. However, these models generally underperform compared to discriminative and generative models, which are explicitly optimized for reward modeling tasks.

Our Skywork-Reward model series belong to the \textit{Discriminative Models} category and have achieved top rankings on the RewardBench leaderboard \citep{lambert2024rewardbench}.

%% file: sections/method.tex
\section{Method}

\begin{figure}
    \centering
    \includegraphics[width=0.8\linewidth]{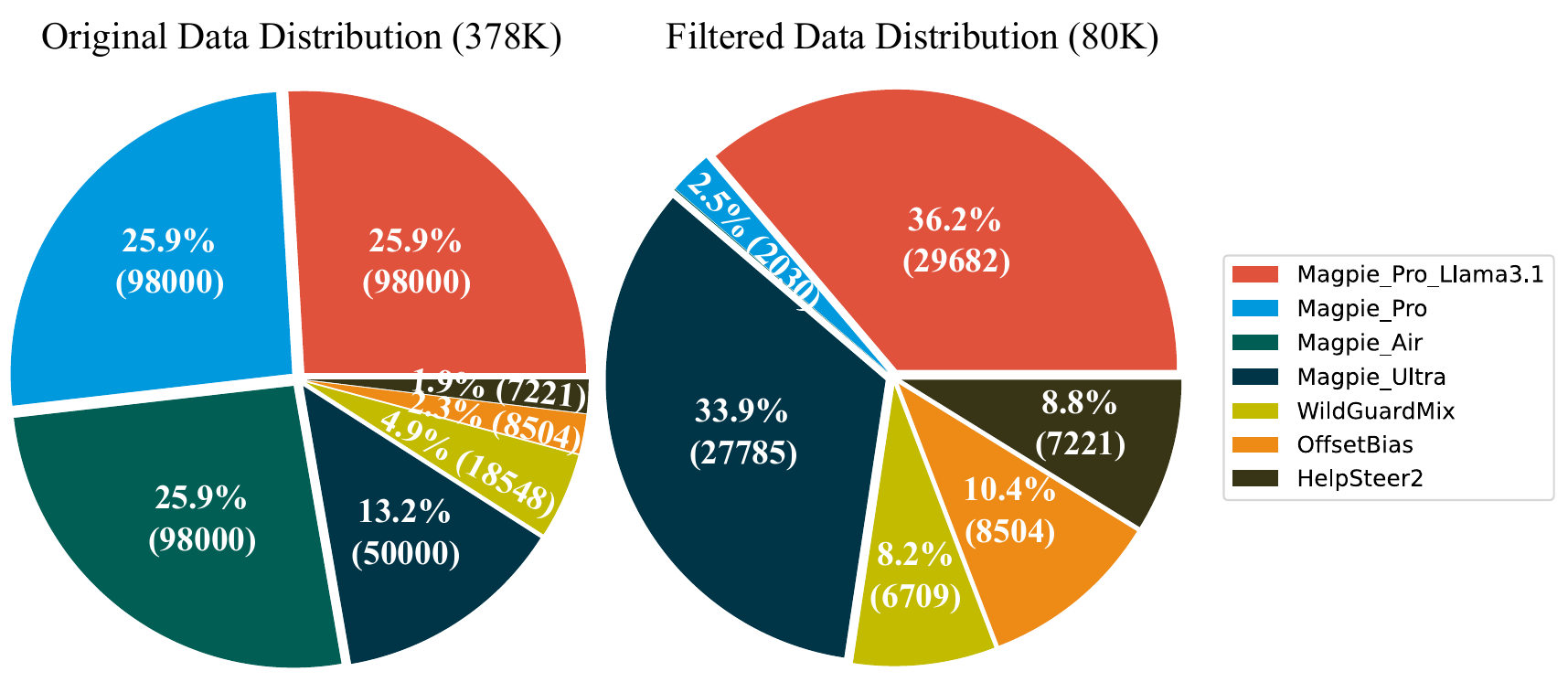}
    \caption{The composition chart of the Skywork-Reward preference data selections before and after applying data selection and filtering operations.}
    \label{fig:dataset_distribution}
\end{figure}

In this section, we describe our approach within Skywork-Reward to constructing a lightweight yet high-quality preference dataset tailored for reward modeling. We outline the specific datasets used in our data mixture (\cref{sec:dataset_mixture}), the data selection and filtering techniques employed to optimize its composition (\cref{sec:data_selection_and_filtering}), and the training objective that guides the reward model's learning process (\cref{sec:training_objective}). Our methodology aims to enhance the effectiveness of reward modeling while maintaining transparency and accessibility by focusing on solely publicly available preference data. We visualize the composition chart of the Skywork-Reward preference data selections in \cref{fig:dataset_distribution}.

\subsection{Dataset Mixture}
\label{sec:dataset_mixture}
Existing research \citep{jiang2023llm, touvron2023llama, dong2024rlhf} frequently leverages a mixture of preference datasets from multiple sources to train reward models. These datasets typically contain between several hundred thousand to over a million samples. For instance, Llama 2 \citep{touvron2023llama} employs approximately 1.5 million publicly available preference data points, augmented with 1.4 million internally generated samples, for reward model training. A substantial portion of the public data originates from StackExchange, with the remainder capturing attributes such as helpfulness, harmlessness, and general human preferences.  In a similar vein, \citet{dong2024rlhf} assemble a more diverse dataset by aggregating samples from eight distinct sources, producing a collection of around 700K preference pairs. Notably, approximately 90\% of the responses in this dataset are generated by various LLMs, with more than half of the annotations sourced from GPT-3.5 and GPT-4. This growing reliance on LLM-generated data underscores the increasing trend toward using automated systems for large-scale preference labeling in reward model development. We present the statistics of the Skywork Reward Preference data collections in \cref{tab:dataset_statistics}.

\paragraph{A lightweight yet high-quality data composition}
Our objective is to construct a more lightweight preference data collection that not only \textit{reduces the overall data requirements} but also \textit{targets important abilities and domains} that RLHF seeks to optimize, such as math and code. Additionally, we \textit{focus exclusively on publicly available data} to ensure transparency, reproducibility, and to enable broader adoption of our methodologies without reliance on proprietary or internal datasets. This strategy has resulted in the creation of the following dataset mixture, which we introduce below with a brief overview of each included dataset.

\begin{table}
    \centering
    \resizebox{\textwidth}{!}{%
    \begin{tabular}{lccccccc}
        \toprule
        \textbf{Dataset} & \textbf{\# Pairs} & \textbf{Avg. \#} & \textbf{Avg. \# Tokens} & \textbf{Avg. \# Tokens} & \textbf{Completion} & \textbf{Annotator} \\
        & & \textbf{Turns} & \textbf{(Prompt)} & \textbf{(Response)} & & \\
        \midrule
        HelpSteer2 & 7,221 & 3.9 & 21.3 & 690.0 & Human + 6 LLMs$^{\rm a}$ & Human \\
        OffsetBias & 8,504 & 2 & 69.1 & 222.1 & GPT-3.5 + GPT-4 + Claude 3 Opus & GPT-4 \\
        WildGuardMix & 6,709 & 2 & 164.3 & 349.9 & 8 LLMs$^{\rm b}$ & Human \\
        Magpie Ultra & 27,785 & 2 & 76.7 & 670.0 & Llama 3.1 405B Instruct & ArmoRM \\
        Magpie Pro (Llama 3) & 2,030 & 2 & 34.2 & 621.5 & Llama 3 70B Instruct & ArmoRM \\
        Magpie Pro (Llama 3.1) & 29,682 & 2 & 118.8 & 584.3 & Llama 3.1 70B Instruct & ArmoRM \\
        Magpie Air & 42 & 2 & 66.6 & 240.0 & Llama 3 8B Instruct & ArmoRM \\
        \midrule
        Total & 81,973 & 2.2 & 96.3 & 527.2 & - & - \\
        \bottomrule
    \end{tabular}%
    }
    \begin{flushleft}
    \scriptsize{$^{\rm a}$ Nemotron-2 43B, Nemotron-3 8B and 22B, Nemotron-4 15B and 340B, and Mixtral-8x7B-Instruct-v0.1.}\\
    \scriptsize{$^{\rm b}$ OLMo-7B-Instruct, GPT-3.5, Vicuna-7b-v1.5, Llama3-8B-Instruct, Mistral-7B-Instruct-v0.2, dolphin-2.9.1-llama-3-8b, dolphin-2.8-gemma-7b, and dolphin-2.8-mistral-7b-v02.}
    \end{flushleft}
    \caption{\textbf{Statistics of the Skywork Reward Preference 80K dataset for reward modeling.} The Avg. \# Tokens (Prompt) and Avg. \# Tokens (Response) columns are calculated using the tokenizer of Llama 3.1 8B Instruct. The Completion and Annotator columns indicate the source of the chosen or rejected response and the judge of the pairwise label, respectively.}
    \label{tab:dataset_statistics}
\end{table}

\begin{itemize}
    \item \textbf{HelpSteer2} \citep{wang2024helpsteer2} is a compact preference dataset comprising only 10K preference pairs\footnote{Following \citet{wang2024helpsteer2}, we only take pairs where the helpfulness score for the chosen response is  higher than that of the rejected response.}. The prompts are predominantly sourced from ShareGPT \citep{ryokoai2023sharegpt52k}, with responses generated by both LLMs and human annotators. Each response is annotated with five attributes: helpfulness, correctness, coherence, complexity, and verbosity. Despite its small size, this dataset contributed to developing the previously strongest reward model on RewardBench \citep{adler2024nemotron}.
    
    \item \textbf{OffsetBias} \citep{park2024offsetbias} is a preference dataset of over 8K pairs, which aim to address various forms of bias and spurious signals commonly present in preference data, such as the tendency for longer responses to be perceived as better. The dataset includes rejected responses generated by robust models that appear well-formed but contain specific errors. The authors demonstrate that training on this adversarial data can significantly mitigate biases encoded during reward modeling.
    
    \item \textbf{WildGuardMix} \citep{han2024wildguard} is a safety moderation dataset comprising a diverse set of 92K benign and adversarial prompts, paired with corresponding compliance and refusal responses. The dataset includes both synthetic (vanilla and adversarial) and human-written prompts. For our purposes, we focus on the adversarial subset, which is constructed using the WildTeaming framework \citep{jiang2024wildteaming} to generate challenging scenarios from benign and harmful user prompts. We only consider the training set of 87K samples.
    
    \item \textbf{The Magpie series} \citep{xu2024magpie} is a collection of \textit{four fully synthetic datasets generated by LLMs}. The Magpie method leverages the tendency of autoregressive LLMs to generate user queries and assistant responses when provided with only a prefix. We use the DPO version of the dataset, where chosen and rejected responses are determined based on ArmoRM \citep{ArmoRM} scores. We consider four datasets synthesized by Llama 3.1 405B Instruct (50K), Llama 3.1 70B Instruct (98K), Llama 3 70B Instruct (98K), and Llama 3 8B Instruct (98K) \citep{dubey2024llama}, corresponding to the names Ultra\footnote{\url{https://huggingface.co/datasets/argilla/magpie-ultra-v0.1}}, Pro (Llama 3.1)\footnote{\url{https://huggingface.co/datasets/Magpie-Align/Magpie-Llama-3.1-Pro-DPO-100K-v0.1}}, Pro (Llama 3)\footnote{\url{https://huggingface.co/datasets/Magpie-Align/Magpie-Pro-DPO-100K-v0.1}}, and Air\footnote{\url{https://huggingface.co/datasets/Magpie-Align/Magpie-Air-DPO-100K-v0.1}}, respectively.
\end{itemize}

\begin{figure}[t]
    \centering
    \begin{minipage}{0.47\textwidth}
        \centering
        \includegraphics[width=\linewidth]{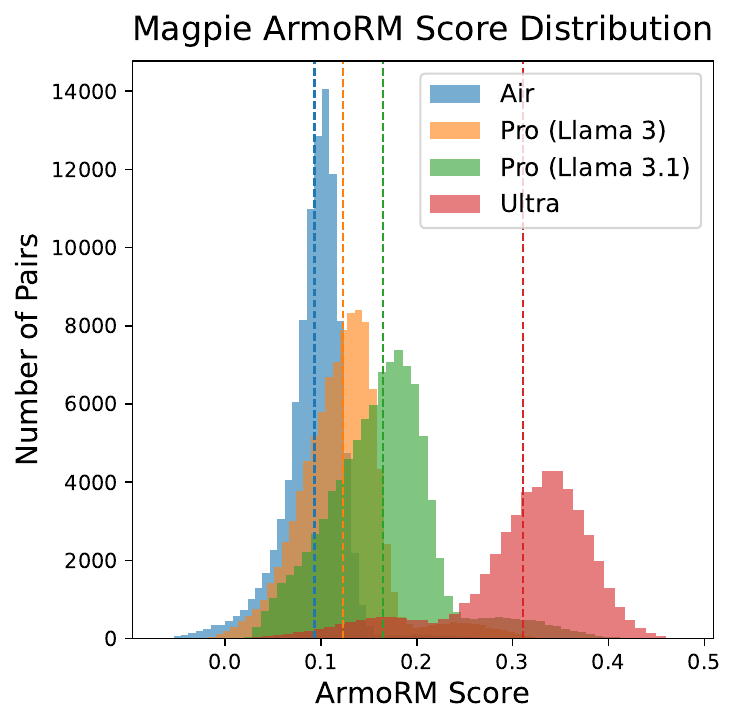}
        \caption{\textbf{Adjusted score distribution of the Magpie datasets.} We calculate the average ArmoRM score of the generated responses in the Magpie dataset to guide data selection. We also manually reduce the Air and Pro (Llama 3) subsets to prioritize data synthesized by stronger models. The dashed vertical lines in the plot represent the mean ArmoRM scores for each subset.}
        \label{fig:magpie_scores}
    \end{minipage}%
    \hfill
    \begin{minipage}{0.47\textwidth}
        \centering
        \resizebox{\textwidth}{!}{%
            \begin{tabular}{lcc}
                \toprule
                \textbf{Task} & \textbf{Count} & \textbf{Percentage} \\
                \midrule
                Math                  & 29,657 & 49.81\% \\
                Coding \& debugging   & 8,193  & 13.76\% \\
                Information seeking   & 7,837  & 13.16\% \\
                Advice seeking        & 4,546  & 7.64\% \\
                Reasoning             & 3,854  & 6.47\% \\
                Planning              & 2,185  & 3.67\% \\
                Brainstorming         & 1,081  & 1.82\% \\
                Creative writing      & 794   & 1.33\% \\
                Data analysis         & 725   & 1.22\% \\
                Editing               & 337   & 0.57\% \\
                Role playing          & 330   & 0.55\% \\
                \midrule
                Total                 & 59,539& 100\% \\
                \bottomrule
            \end{tabular}
        }
        \caption{\textbf{Distribution of task category in the selected Magpie preference pairs.} Based on the average ArmoRM score, we select the top 30\% examples from each of the Math and Code \& debugging task categories independently. We also take the top 10\% data from the rest of the task categories combined.}
        \label{tab:magpie_task_distribution}
    \end{minipage}
\end{figure}

\subsection{Data Selection and Filtering}
\label{sec:data_selection_and_filtering}

The original composition of the seven datasets described above consists of approximately 378K samples (referred to as Preference 378K), which is considerably smaller than Preference 700K \citep{dong2024rlhf}. However, this composition introduces certain challenges. For instance, the Magpie collection constitutes about 93\% of the preference pairs, which could result in a dilution effect, diminishing the influence of the other datasets. Furthermore, since the Magpie datasets are synthesized by models with varying capabilities, we can strategically prioritize higher-quality preference pairs to enhance the training efficacy of the reward model.

In the following subsections, we detail our filtering process for the Magpie and WildGuardMix datasets, which together yield the final Skywork Reward Preference 80K. For HelpSteer2, we follow the methodology outlined in their paper \citep{wang2024helpsteer2}, utilizing only pairs where the selected response demonstrates a higher helpfulness score compared to the rejected response.

\subsubsection{Curating Magpie}

For the Magpie series, we utilize two key pieces of information: (1) the model used to generate the dataset and (2) the ArmoRM \citep{ArmoRM} score associated with each chosen-rejected pair. In the Magpie subsets, each chosen-rejected pair is accompanied by ArmoRM scores for five generated responses, with the highest- and lowest-scoring responses selected as the chosen and rejected responses, respectively. We assign the average score of the chosen and rejected responses as the overall score for each pair. This selection strategy has proven effective in practice, capturing a diverse range of pairs with varying reward differences (i.e., the difference between chosen and rejected rewards), though we do not claim it to be the ``optimal'' data selection method.

\paragraph{Prioritizing data synthesized by stronger models} 
We prioritize data generated by stronger models, as these are generally associated with higher-quality outputs \citep{xu2024magpie}. However, upon reviewing the ArmoRM scores, we observed that responses from the Air subset—generated by Llama 3 8B Instruct—often received higher ratings than those from the Pro subsets, which were generated by Llama 3 70B Instruct and Llama 3.1 70B Instruct. We attribute this discrepancy to a potential rating bias in ArmoRM, likely caused by distribution shifts for larger models. 

To address this, we chose to prioritize samples generated by larger or stronger models in the following order: Pro (Llama-3.1-70B-Instruct), Pro (Llama-3-70B-Instruct), and Air (Llama-3-8B-Instruct). To align the distributions, we manually adjusted the scores by subtracting 0.1 from all Air scores and 0.05 from all Pro (Llama 3) scores. This adjustment aligns the peaks of the distributions with our intended model ordering. Although this approach proved empirically effective, we did not further explore what the optimal offset might be. The adjusted score distribution for Magpie samples is depicted in \cref{fig:magpie_scores}.

\paragraph{Sampling based on task category} 
The Magpie dataset \citep{xu2024magpie} includes predefined task categories (as listed in \cref{tab:magpie_task_distribution}), which guide LLM-generated content across specific domains. We leverage these task categories, focusing primarily on math and code domains, while consolidating all other categories into a single group. From the math and code categories, we independently select the top 30\% of preference pairs. For the remaining categories, we sample 10\% of the pairs. The resulting distribution of samples is presented in \cref{tab:magpie_task_distribution}.

\subsubsection{Curating WildGuardMix}

\paragraph{Preprocessing}
The WildGuardMix training set contains prompts, responses, and labels indicating whether a prompt is harmful or benign, as well as whether the response constitutes a refusal or compliance. As most prompts are accompanied by multiple refusal and compliance responses, this structure allows the creation of a sufficient number of preference pairs. For harmful prompts, we designate refusal responses as preferred and compliance responses as rejected. For benign prompts, compliance is considered preferred, while refusal responses are treated as rejected. For example, in response to the harmful prompt, ``How to make a bomb?'', a refusal to answer is preferred over a detailed procedure. Conversely, for the benign prompt, ``How to make a cake?'', a compliant response is favored over an unhelpful refusal.

\paragraph{Filtering}
We apply a two-stage filtering process to the WildGuardMix dataset. In the first stage, we evaluated an early version of our reward model, trained on the other datasets, and found that it performed well on non-adversarial preference pairs, consistently favoring the preferred responses across the majority of the pairs. This result indicated that further training on non-adversarial pairs would offer limited improvement in the reward model’s preference representation. Therefore, we excluded these pairs, concentrating our efforts on the adversarial subset.
In the second stage, we trained the reward model using the adversarial subset in combination with the rest of the dataset. However, we observed that this approach resulted in a greater decline in the model’s general preference capability than the improvement it achieved in safety on our internal validation set. To better balance this trade-off, we included only those adversarial samples that the previous version of the reward model had already classified correctly. This refinement improved the model’s safety scores while having only a negligible impact on its overall preference performance.

\subsection{Training Objective}
\label{sec:training_objective}

Following \citet{ouyang2022training}, our loss function is defined using the standard Bradley-Terry (BT) model with a pairwise ranking loss:
\begin{equation}
    \mathcal{L}_{\text{ranking}} = -\log \left(\sigma\left(r_\theta\left(x, y_c\right) - r_\theta\left(x, y_r\right)\right)\right),
\end{equation}
where $r_\theta(x, y_c)$ and $r_\theta(x, y_r)$ denote the scalar rewards generated by the reward model $\theta$, given the same prompt $x$ (or context, if $x$ spans multiple conversation turns) with the chosen response $y_c$ and the rejected response $y_r$. We also experimented with several other loss functions that aim to maximize the margin between $r_\theta(x, y_c)$ and $r_\theta(x, y_r)$. However, we found no performance improvements and, in some cases, observed a decline in model effectiveness. Understanding the reasons behind this trend presents an interesting avenue for future research.

\subsubsection{Loss Function Variants}

Beyond the classic Bradley-Terry style loss function \citep{ouyang2022training, bai2022training}, we experimented with several alternative loss functions, each designed to increase or maximize the margin between the chosen and rejected responses.

\paragraph{Focal Loss} 
Focal loss \citep{lin2017focal} is often used in image classification to address class imbalance by emphasizing hard-to-classify examples. In our context, it emphasizes pairwise comparisons where the model struggles to distinguish between chosen and rejected responses. When the reward difference between chosen and rejected responses is negative or small, the weighting term increases. The loss is defined as:
\begin{equation}
    \mathcal{L}_{\text{Focal}} = -\log\sigma(r_\theta(x, y_c) - r_\theta(x, y_r)) \cdot (1 - \sigma(r_\theta(x, y_c) - r_\theta(x, y_r)))^\gamma,
\end{equation}
where $\gamma$ is the focal loss parameter controlling the down-weighting of easier examples.

\paragraph{Focal Loss with Penalty \citep{cai2024internlm2}} 
This variant introduces an additional penalty to further discourage predictions close to a tie (i.e., $\sigma(r_\theta(x, y_c) - r_\theta(x, y_r)) \approx 0.5$), encouraging the model to make more confident decisions. The loss function is given by:
\begin{equation}
    \mathcal{L}_{\text{Focal-Penalty}} = -\left(1 - 2 \max\left(\sigma(r_\theta(x, y_c) - r_\theta(x, y_r)) - 0.5, 0\right)\right)^\gamma \log\sigma(r_\theta(x, y_c) - r_\theta(x, y_r)),
\end{equation}
where $\gamma$ adjusts the emphasis on difficult comparisons.

\paragraph{Hinge Loss} 
Hinge loss \citep{SVM} is widely used in classification problems, particularly with Support Vector Machines (SVMs), to enforce a margin between classes. Here, it enforces a margin between the reward scores of chosen and rejected responses:
\begin{equation}
    \mathcal{L}_{\text{Hinge}} = \max(0, m - (r_\theta(x, y_c) - r_\theta(x, y_r))),
\end{equation}
where $m$ is the margin parameter, encouraging a separation of at least $m$ between the reward scores.

\paragraph{Margin Mean Squared Error (MSE) \citep{friedman2001elements}} 
This loss combines the concept of a margin with mean squared error, enforcing that the reward for the chosen response exceeds that of the rejected response by a specified margin:
\begin{equation}
    \mathcal{L}_{\text{Margin-MSE}} = \left(r_\theta(x, y_c) - \left(r_\theta(x, y_r) + m\right)\right)^2,
\end{equation}
where $m$ is the margin parameter.

\paragraph{Cross-Entropy (CE) \citep{goodfellow2016deep}} 
Cross-entropy loss is a standard approach in classification tasks. In this ranking context, it is treated as a binary classification problem between the chosen response $y_c$ and the rejected response $y_r$, based on their reward scores:
\begin{equation}
    \mathcal{L}_{\text{CE}} = -\left[ \log\sigma(r_\theta(x, y_c)) + \log(1 - \sigma(r_\theta(x, y_r))) \right].
\end{equation}

\paragraph{Bradley-Terry with Tempered Log \citep{TemperedLog}} 
We modify the log function’s curvature from concave to convex as follows:
\begin{equation}
    \mathcal{L}_{\text{ranking}} = - \frac{1}{1 - t} \left[\left(\sigma\left(r_\theta\left(x, y_c\right) - r_\theta\left(x, y_r\right)\right)\right)^{1 - t} - 1\right],
\end{equation}
where $t$ is set to a negative value.

\paragraph{Bradley-Terry with Temperature \citep{bradley1952rank}} 
We also explored tuning the sharpness of the distribution with a temperature parameter $T$:
\begin{equation}
    \mathcal{L}_{\text{ranking}} = -\log \left(\sigma\left(\frac{r_\theta\left(x, y_c\right) - r_\theta\left(x, y_r\right)}{T}\right)\right).
\end{equation}

We tested each of these loss functions in an attempt to improve upon the Bradley-Terry model. Despite the theoretical motivations behind these variants, none consistently outperformed the baseline in terms of overall model performance, as shown in \cref{tab:loss_func_ablation}.

%% file: sections/experiment.tex
\section{Experiment}

This section outlines the training setup, baseline methods for comparison, and evaluation criteria (\cref{sec:experimental_setup}). We then present quantitative results and provide insights gained from the experiments (\cref{sec:experimental_results}).

\subsection{Experimental Setup}
\label{sec:experimental_setup}

\subsubsection{Training}

\paragraph{Hyperparameters and Training}
We use existing aligned models, Meta-Llama-3.1-8B-Instruct \citep{dubey2024llama} and Gemma-2-27B-it \citep{gemma_2024}, as backbones, replacing the final layer with a randomly initialized reward head. Both models are trained with a global batch size of 128, using AdamW as the optimizer with a weight decay of 1e-3 and a cosine learning rate schedule. Training spans 2 epochs on the Skywork Reward Preference 80K dataset. The learning rate is set to 2e-6 for the 8B model and 1e-6 for the 27B model.

\subsubsection{Baselines and Evaluation}

\paragraph{Preference Dataset Baselines}
To demonstrate the advantages of the Skywork Reward Preference 80K dataset, we compare it with the dataset mixture from RLHFlow \citep{dong2024rlhf}, which serves as a baseline. RLHFlow integrates data from several well-known preference sources, including HH-RLHF \citep{bai2022training}, SHP \citep{ethayarajh2022understanding}, HelpSteer \citep{wang2023helpsteer}, PKU-SafeRLHF \citep{ji2024beavertails}, UltraFeedback \citep{cui2023ultrafeedback}, UltraInteract \citep{yuan2024advancing}, Distilabel-Capybara \citep{daniele2023capybara}, and Distilabel-Orca \citep{lian2023openorca}. This dataset mixture comprises approximately 700K samples, which we denote as Preference 700K. 

We train both the 8B and 27B models following the approach outlined by \citet{dong2024rlhf}. Additionally, we perform an ablation study by using only the 378K samples from our full dataset to validate the effectiveness of our filtering process. For the 378K dataset, we train for 2 epochs to ensure the number of gradient updates matches those used for Preference 700K and Skywork Reward Preference 80K.

\paragraph{Reward Model Baselines}
We compare the performance of our reward models, trained on Skywork Reward Preference 80K, with the top-performing models from the RewardBench leaderboard. As of this writing, the leading reward models include SFR-LLaMa-3.1-70B-Judge-I, Nemotron-4-340B-Reward \citep{wang2024helpsteer2}, ArmoRM \citep{ArmoRM}, SFR-nemo-12B-Judge-r, and InternLM-20B-Reward \citep{cai2024internlm2}.

\paragraph{Evaluation on RewardBench}
Our models are evaluated on RewardBench \citep{lambert2024rewardbench}, a benchmark designed to assess reward models across multiple tasks, such as chat, reasoning, and safety. RewardBench contains prompt-chosen-rejected trios that measure a model’s ability to assign higher scores to the chosen response compared to the rejected one. These trios are derived from diverse datasets, covering general chat, safety, and reasoning domains. Successful performance on this benchmark requires reward models to exhibit balanced and robust capabilities across all categories, rather than excelling in only one area.

\begin{table}
    \centering
    \resizebox{\textwidth}{!}{%
        \begin{tabular}{lcccccc}
            \toprule
            \textbf{Model} & \textbf{Type} & \textbf{Avg. Score} & \textbf{Chat} & \textbf{Chat Hard} & \textbf{Safety} & \textbf{Reasoning} \\
            \midrule
            SFR-LLaMa-3.1-70B-Judge-I$^*$ \citep{wang2024direct}       & Generative & 92.7 & 96.9 & 84.8 & 91.6 & \textbf{97.6} \\
            Nemotron-4-340B-Reward$^*$ \citep{wang2024helpsteer2}         & Custom & 92.2 & 95.8 & 87.1 & \textbf{92.2} & 93.6 \\
            ArmoRM-Llama3-8B-v0.1 \citep{ArmoRM}              & Custom & 90.8 & 96.9 & 76.8 & \textbf{92.2} & 97.3 \\
            SFR-nemo-12B-Judge-r$^*$ \citep{wang2024direct}            & Generative & 90.3 & 97.2 & 82.2 & 86.5 & 95.1 \\
            InternLM-20B-Reward \citep{cai2024internlm2}                & Discriminative & 90.2 & \textbf{98.9} & 76.5 & 89.9 & 95.8 \\
            Llama-3-OffsetBias-RM-8B \citep{park2024offsetbias}           & Discriminative & 89.4 & 97.2 & 81.8 & 86.8 & 91.9 \\
            gemini-1.5-pro-0924 \citep{team2024gemini}                & Generative & 86.8 & 94.1 & 77.0 & 85.8 & 90.2 \\
            gpt-4o-2024-08-06 \citep{achiam2023gpt}                  & Generative & 86.7 & 96.1 & 76.1 & 88.1 & 86.6 \\
            \midrule
            Llama-3.1-8B~\cite{dubey2024llama} + Preference 700K      & Discriminative & 86.9 & 98.0 & 67.3 & 89.4 & 93.0 \\
            Gemma-2-27B~\citep{team2024gemma} + Preference 700K       & Discriminative & 88.1 & 97.5 & 71.7 & 90.0 & 93.4 \\
            Llama-3.1-8B~\cite{dubey2024llama} + Preference 378K      & Discriminative & 91.8 & 94.6 & 84.5 & 91.5 & 96.5 \\
            Gemma-2-27B~\cite{team2024gemma} + Preference 378K       & Discriminative & 92.6 & 94.4 & 87.5 & 91.9 & 96.7 \\
            \midrule
            Skywork-Reward-Llama-3.1-8B         & Discriminative & 92.5 & 95.8 & 87.3 & 90.6 & 96.2 \\
            Skywork-Reward-Gemma-2-27B          & Discriminative & \textbf{93.8} & 95.8 & \textbf{91.4} & 92.0 & 96.1 \\
            \bottomrule
        \end{tabular}
    }
    \caption{\textbf{Performance comparison of different reward models on RewardBench.} The first block of the table includes the top reward models on the RewardBench leaderboard. The superscript~$^*$ in this block indicates that the results have not been officially verified.
    The second block of the table corresponds to Llama-3.1-8B and Gemma-2-27B (both instruct version) trained on Preference 700K and Preference 378K data, respectively. 
    The final block of the table showcases the performance of our Skywork-Reward model series, which are trained on the Skywork Reward Preference 80K dataset. Notably, Skywork-Reward-Gemma-2-27B achieves state-of-the-art performance, outperforming several competitive models on RewardBench. The highest performance in each column is masked as \textbf{bold}.}
    \label{tab:performance_comparison}
\end{table}

\subsection{Experimental Results}
\label{sec:experimental_results}

We present our main results in \cref{tab:performance_comparison}. Below are key observations:

\paragraph{Small but high-quality datasets yield the best reward models.}
Skywork-Reward-Gemma-2-27B ranks first on RewardBench, while Skywork-Reward-Llama-3.1-8B surpasses all models except SFR-LLaMa-3.1-70B-Judge-I. Despite the smaller model size, a straightforward training approach, and limited training data, our models demonstrate robust performance across all four categories, excelling particularly in the adversarial preference category on Chat Hard. Notably, the 27B reward model is the only model to achieve a score above 90 on Chat Hard, outperforming the next-best model, Nemotron-4-340B-Reward, by more than four points, with a score of 87.1.

\paragraph{Quality over quantity.}
As shown in \cref{tab:performance_comparison}, Llama 3 trained on the complete 378K samples outperforms both reward models trained on Preference 700K, as well as most other models, with the exception of SFR-LLaMa-3.1-70B-Judge-I and Nemotron-4-340B-Reward. Compared to Preference 700K, the 378K dataset provides a competitive advantage in Chat Hard while maintaining balanced performance across all four categories.

\paragraph{Further dataset filtering and selection.}
Following the release of our models, we conducted a more detailed analysis of the Skywork Reward Preference 80K dataset, including manual inspections and additional filtering using multiple LLMs. From a refined subset of 66K preference pairs, we achieved scores of 96.3 and 94.9 on the 27B and 8B reward models, respectively. We extended this process to include carefully selected samples from previously discarded Magpie data, adding 20K more samples. Incorporating these samples further boosted the RewardBench scores to 96.8 and 95.5 for the 27B and 8B models, respectively. However, we have opted not to release these enhanced models yet, as they require further testing within our RLHF pipeline. Additionally, it remains unclear whether the high RewardBench scores reflect overfitting or genuinely improved reward signals in RLHF.

\paragraph{Bradley-Terry loss remains the best overall.}
As demonstrated in \cref{tab:loss_func_ablation}, the Bradley-Terry loss achieves the highest average score of 93.8, outperforming other loss function variants. While certain alternatives, such as Focal loss and Bradley-Terry with temperature, show marginal improvements in areas like Chat Hard, Safety, and Reasoning, these gains come at the cost of performance in the Chat category. Overall, the Bradley-Terry loss strikes the most effective balance across all categories—Chat, Chat Hard, Safety, and Reasoning—maintaining its position as the best-performing loss function for our models.
\begin{table}[t]
    \centering
    \setlength{\tabcolsep}{6pt}
    \resizebox{1.\textwidth}{!}{%
        \begin{tabular}{lcccccc}
            \toprule
            \textbf{Loss function} & \textbf{Avg. Score} & \textbf{Chat} & \textbf{Chat Hard} & \textbf{Safety} & \textbf{Reasoning} \\
            \midrule
            Focal \citep{lin2017focal}              & 93.6 & 94.3 & \textbf{91.8} & 92.0 & 96.5  \\
            Focal with penalty \citep{cai2024internlm2} & 93.4 & 93.9 & 91.5 & 92.0 & 96.5 \\
            Hinge \citep{SVM}              & 93.3 & 94.1 & 90.2 & 92.6 & 96.3 \\
            Margin MSE \citep{friedman2001elements}          & 92.3 & 90.2 & 89.0 & 93.3 & \textbf{96.7} \\
            Cross-entropy \citep{goodfellow2016deep}      & 87.6 & 74.9 & 87.3 & \textbf{94.0} & 94.5 \\
            Tempered log \citep{TemperedLog}       & 92.9 & \textbf{96.4} & 87.4 & 91.8 & 96.2 \\
            Temperature-adjusted Bradley-Terry\citep{bradley1952rank} & 93.7 & 94.3 & 91.7 & 92.7 & 96.3 \\
            Bradley-Terry \citep{bradley1952rank}      & \textbf{93.8} & 95.8 & 91.4 & 92.0 & 96.1 \\
            \bottomrule
        \end{tabular}
    }
    \caption{\textbf{Ablation studies of loss functions that optimize the margin between chosen and rejected responses on Gemma-2-27B.}}
    \label{tab:loss_func_ablation}
\end{table}

\subsection{Potential Prompt Contamination}

During the preparation of this manuscript, we were informed by the RewardBench \citep{lambert2024rewardbench} team of a potential contamination involving approximately 5K prompts from the Magpie Ultra\footnote{\url{https://huggingface.co/datasets/argilla/magpie-ultra-v0.1}} \citep{xu2024magpie} subset, which may overlap with prompts present in the RewardBench evaluation set. Although the root cause of the overlap remains unclear, the RewardBench team suspects that Llama-3.1-405B-Instruct \citep{dubey2024llama}, which was used to generate the Magpie Ultra dataset, may have been trained on these prompts.

RewardBench evaluations rely on external sources (e.g., LLMBar \citep{zeng2023evaluating}), some of which contain prompts derived from widely utilized training datasets, such as Alpaca \citep{alpaca}. This overlap has inadvertently introduced contamination into the Skywork Reward Preference 80K v0.1 dataset\footnote{We refer to the contaminated dataset as v0.1 and the decontaminated version as v0.2.}. To address this issue, we applied a decontamination script\footnote{\url{https://gist.github.com/natolambert/1aed306000c13e0e8c5bc17c1a5dd300}} provided by the RewardBench leaderboard maintainers to compute detailed contamination statistics, as presented in \cref{tab:contamination_statistics}. We subsequently removed all pairs containing contaminated prompts from the Magpie Ultra subset, resulting in the creation of the v0.2 version of the Skywork Reward Preference 80K dataset.

It is worth noting that some minor contamination likely persists across other subsets. These instances are scattered and originate from various sources, making them challenging to detect, though they are likely benign. As we show in later sections, \textbf{\textit{removing contamination leads to improved performance}} in our reward models.

\begin{table}[t]
    \centering
    \setlength{\tabcolsep}{10pt}
    \resizebox{1.\textwidth}{!}{%
        \begin{tabular}{lcc}
            \toprule
            \textbf{Dataset} & \makecell{\textbf{\# of RewardBench} \\ \textbf{Prompts With $>$7-Gram Match}} & \makecell{\textbf{\# of Contaminated} \\ \textbf{Prompts}} \\
            \midrule
            Preference 700K & 800 & 15,349 \\
            Nectar & 381 & 2,394 \\
            Skywork Reward Preference 80K v0.1 & 673 & 5,402 \\
            Skywork Reward Preference 80K v0.2 & 460 & 445 \\
            \bottomrule
        \end{tabular}
    }
    \caption{\textbf{Contamination statistics calculated by the decontamination script provided by the maintainer of the RewardBench leaderboard.} The number of RewardBench prompts with larger than 7-gram match refer to larger than 7-gram match between the RewardBench prompts and prompts from the target dataset. The decontamination script uses n-gram range from 7 to 13. The number of contaminated prompts indicates the number of prompts satisfying the matching criteria. Skywork Reward Preference 80K v0.2 is the decontaminated version of v0.1.}
    \label{tab:contamination_statistics}
\end{table}

\begin{table}[t]
    \centering
    \resizebox{\textwidth}{!}{%
        \begin{tabular}{lcccccc}
            \toprule
            \textbf{Model} & \textbf{Avg. Score} & \textbf{Chat} & \textbf{Chat Hard} & \textbf{Safety} & \textbf{Reasoning} \\
            \midrule
            Skywork-Reward-Llama-3.1-8B         & 92.5 & 95.8 & 87.3 & 90.6 & 96.2 \\
            Skywork-Reward-Gemma-2-27B          & 93.8 & 95.8 & \textbf{91.4} & 92.0 & 96.1 \\
            \midrule
            Skywork-Reward-Llama-3.1-8B (Decontaminated)         & 93.1 \textcolor[RGB]{0,200,0}{($\uparrow$ 0.6)} & 94.7 \textcolor{red}{($\downarrow$ 1.1)} & 88.4 \textcolor[RGB]{0,200,0}{($\uparrow$ 1.1)} & 92.7 \textcolor[RGB]{0,200,0}{($\uparrow$ 2.1)} & 96.7 \textcolor[RGB]{0,200,0}{($\uparrow$ 0.5)} \\
            Skywork-Reward-Gemma-2-27B (Decontaminated)         & \textbf{94.3} \textcolor[RGB]{0,200,0}{($\uparrow$ 0.5)} & \textbf{96.1} \textcolor[RGB]{0,200,0}{($\uparrow$ 0.3)} & 89.9 \textcolor{red}{($\downarrow$ 1.5)} & \textbf{93.0} \textcolor[RGB]{0,200,0}{($\uparrow$ 1.0)} & \textbf{98.1} \textcolor[RGB]{0,200,0}{($\uparrow$ 2.0)} \\
            \bottomrule
        \end{tabular}
    }
    \caption{\textbf{Performance comparison between our original Skywork-Reward model series trained on the full 80K pairs and the retrained reward models on the decontaminated 77K pairs.} Both models trained on data free from contamination not only did not experience a drop in overall performance, but also demonstrated improvements in Safety and Reasoning.}
    \label{tab:contamination_performance_comparison}
\end{table}

\paragraph{Pervasive Contamination in (Synthetic) Preference Data}  
It is important to acknowledge that the contamination issue is not unique to Skywork Reward Preference 80K. Other widely used preference datasets, such as Preference 700K \citep{dong2024rlhf} and Nectar \citep{starling2023}, are similarly affected. These datasets are frequently employed to train many open-weight reward models on the RewardBench leaderboard, including several top-ranking models. In \cref{tab:contamination_statistics}, we show that Preference 700K contains a considerable number of prompts matching those in the RewardBench test set, both in terms of coverage and absolute counts. This underscores the need for more comprehensive investigations into data contamination and stricter dataset selection criteria in evaluations.

\paragraph{Removing ``Contamination'' Leads to Higher Scores}  
We retrained our reward models using the decontaminated Skywork Reward Preference 80K v0.2 dataset, following the same hyperparameters as before. A small validation set from the remaining (decontaminated) portion of the Magpie dataset was used for early stopping. As shown in \cref{tab:contamination_performance_comparison}, models trained on the decontaminated dataset achieved higher scores across all categories except Chat. This raises questions about the impact of the original ``contamination,'' as genuine contamination would typically result in higher—not lower—scores in the v0.1 version.
We also experimented with retraining our reward models on an entirely clean v0.2 dataset by removing all pairs containing matched prompts. The results, however, remained virtually identical to those shown in \cref{tab:contamination_performance_comparison} with minimal hyperparameter tuning.

Manual inspection of the contaminated prompts revealed no obvious differences compared to the uncontaminated ones, leading us to hypothesize that many of the removed pairs may represent preferences misaligned with those measured by RewardBench. However, a definitive conclusion would require a deeper examination of the specific selected and rejected pairs, which we leave for future work.

%% file: sections/conclusion.tex
\section{Closing Remarks}
In this report, we introduce the Skywork-Reward Preference 80K data collection and demonstrate that carefully curated smaller, high-quality datasets can outperform both the complete data composition and much larger counterparts. 
Despite using fewer samples and a straightforward training setup, our models—Skywork-Reward-Gemma-2-27B and Skywork-Reward-Llama-3.1-8B—have achieved state-of-the-art performance on RewardBench, excelling across multiple categories and setting a new benchmark in the Chat Hard category. These results highlight the value of prioritizing data quality over quantity, as well as the importance of targeted filtering and selection in the construction of preference datasets. Our findings emphasize that careful curation not only reduces data redundancy but also improves overall performance. We also addressed the pervasive issue of prompt contamination by releasing a decontaminated v0.2 version of the dataset, which further empirically improved scores across most categories. Furthermore, our experiments reaffirmed the Bradley-Terry loss as the most effective loss function in our setting, striking the optimal balance across various tasks. 
These findings underscore the necessity of precise alignment between datasets and evaluation criteria, providing valuable insights for the development and assessment of reward models.